\documentclass[lettersize,journal]{IEEEtran}
\usepackage{amsmath,amsfonts}
\usepackage{algorithmic}
\usepackage{algorithm}
\usepackage{array}
\usepackage[caption=false,font=normalsize,labelfont=sf,textfont=sf]{subfig}
\usepackage{textcomp}
\usepackage{stfloats}
\usepackage{url}
\usepackage{verbatim}
\usepackage{graphicx}
\usepackage{cite}
\usepackage{amssymb}
\usepackage{multirow}
\usepackage[table]{xcolor}
\begin{document}

\title{Occlusion-Guided Feature Purification Learning via Reinforced Knowledge Distillation for Occluded Person Re-Identification}

\author{Yufei Zheng, Wenjun Wang, Wenjun Gan, Jiawei Liu% <-this % stops a space
% \thanks{This paper was produced by the IEEE Publication Technology Group. They are in Piscataway, NJ.}% <-this % stops a space
% \thanks{Manuscript received April 19, 2021; revised August 16, 2021.}
\thanks{Yufei Zheng, Wenjun Wang, Wenjun Gan, Jiawei Liu are with the Department of Information and Intelligence, University of Science and Technology of China, Hefei 230027, China (Email: zyf2001@mail.ustc.edu.cn; wangwenjun@mail.ustc.edu.cn; ganwenjun@mail.ustc.edu.cn; jwliu6@ustc.edu.cn).}}

% The paper headers
\markboth{}%
{Zheng \MakeLowercase{\textit{et al.}}: OGFR for Occluded Person Re-Identification}

%\IEEEpubid{0000--0000/00\$00.00~\copyright~2021 IEEE}
% Remember, if you use this you must call \IEEEpubidadjcol in the second
% column for its text to clear the IEEEpubid mark.

\maketitle

\begin{abstract}
Occluded person re-identification aims to retrieve holistic images of a given identity based on occluded person images.
Most existing approaches primarily focus on aligning visible body parts using prior information, applying occlusion augmentation to predefined regions, or complementing the missing semantics of occluded body parts with the assistance of holistic images.
Nevertheless, they struggle to generalize across diverse occlusion scenarios that are absent from the training data and often overlook the pervasive issue of feature contamination caused by holistic images. In this work, we propose a novel Occlusion-Guided Feature Purification Learning via Reinforced Knowledge Distillation (OGFR) to address these two issues simultaneously. OGFR adopts a teacher-student distillation architecture that effectively incorporates diverse occlusion patterns into feature representation while transferring the purified discriminative holistic knowledge from the holistic to the occluded branch through reinforced knowledge distillation. Specifically, an Occlusion-Aware Vision Transformer is designed to leverage learnable occlusion pattern embeddings to explicitly model such diverse occlusion types, thereby guiding occlusion-aware robust feature representation. Moreover, we devise a Feature Erasing and Purification Module within the holistic branch, in which an agent is employed to identify low-quality patch tokens of holistic images that contain noisy negative information via deep reinforcement learning, and substitute these patch tokens with learnable embedding tokens to avoid feature contamination and further excavate identity-related discriminative clues. Afterward, with the assistance of knowledge distillation, the student branch effectively absorbs the purified holistic knowledge to precisely learn robust representation regardless of the interference of occlusions. Extensive experiments on the benchmark datasets demonstrate the superiority and effectiveness of OGFR. 
\end{abstract}


\section{Introduction}

\begin{figure}[t]
   %\color{blue}
  \centering
  \includegraphics[width=1.0\columnwidth]{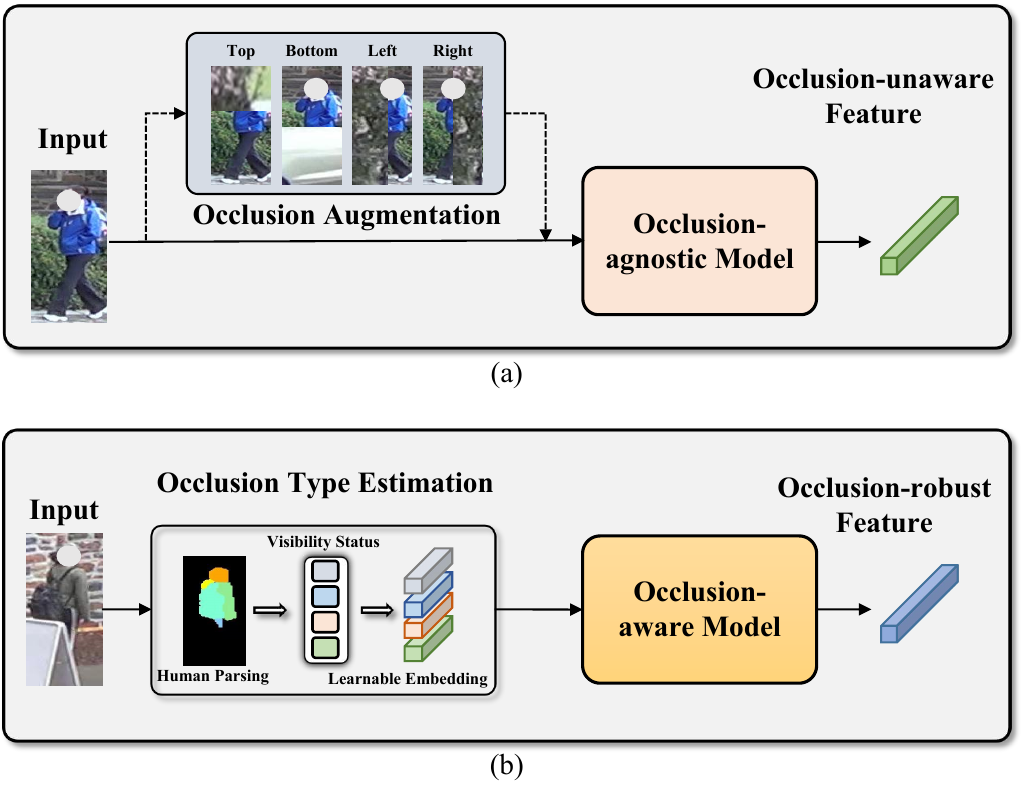}
  \caption{Comparison of OGFR with existing occluded person ReID methods. (a) Existing occlusion-agnostic models either completely neglect occlusion types or rely on occlusion augmentation strategy to passively generate additional occluded images, which are inherently constrained to addressing a limited range of simple occlusions. (b) OGFR proactively incorporates occlusion type cues into learnable occlusion pattern embeddings to effectively handle complex occlusion scenarios and adaptively learn occlusion-aware robust features.}
  \label{fig1}
\end{figure}

\IEEEPARstart{P}{erson} Re-Identification (Re-ID) is the task of identifying the same pedestrian across disjoint camera views, which has extensive applications in video surveillance, smart retail, and automated tracking \textit{etc}. In recent years, significant advancements in person Re-ID \cite{sun2018beyond,he2021transreid,jin2020style,yan2021occluded,ren2022s} have led to impressive performance improvements. However, most existing methods focus on holistic visible images, while overlooking diverse occlusions encountered in real-world scenarios, such as those caused by vehicles, trees, and crowds. Therefore, it is essential to explore effective approaches for occluded person Re-ID. Compared with holistic person Re-ID, occluded person Re-ID is more challenging due to body regions covered by obstacles and other pedestrians, which drastically reduce the availability of discriminative visual cues in feature representation.

To handle above-mentioned challenges, some part-to-part matching methods employ pre-defined rules, \textit{e.g.}, stripe partition strategy \cite{zhang2020semantic,yang2021learning} or introduce external models, such as human parsing \cite{huang2020human} and pose estimation \cite{he2020guided,gao2020pose,miao2021identifying} to learn visible part representations from non-occluded regions. These methods attempt to mitigate occlusion effects by focusing exclusively on visible body parts. 
However, restricting attention solely to non-occluded regions overlooks the impact of different occlusion types on the overall image representation.
As a result, the learned representations become overly dependent on the specific visible regions, leading to degraded performance when faced with diverse occlusions.
An alternative line of research \cite{chen2021occlude, wang2022feature} explores occlusion augmentation techniques to simulate occlusion conditions by artificially generating occluded images. However, these methods typically apply occlusion augmentation to fixed spatial regions, such as occluding the top, bottom, left, or right parts of an image. This results in only four basic occlusion types, which fail to capture the complexity and variability of real-world occlusions. Consequently, these methods exhibit limited generalization capabilities and struggle under diverse and unpredictable occlusion scenarios.
Moreover, these methods provide only the visual content of the image as input, without explicitly incorporating occlusion-specific information.
This omission forces models to infer occlusion patterns purely from visual features, without access to additional contextual cues that could facilitate a more precise understanding. As illustrated in Fig.~\ref{fig1}(a), models that either disregard occlusion types entirely or apply fixed-type occlusion augmentations, both of which lack to incorporate occlusion-related cues in the input, are collectively referred to as occlusion-agnostic models. Such models lack the ability to distinguish between different occlusion types explicitly, making them unstable in feature representation learning and prone to poor generalization across diverse occlusion scenarios.
Alongside the aforementioned approaches, feature recovery methods \cite{zhao2022patch,xu2022learning} which leverage holistic image and visible near-neighbor to complement occluded body parts, have garnered increasing attention in recent years. While these methods aim to restore missing information and improve feature integrity, they still face several limitations. First, the extracted holistic and visible near-neighbor features inherently contain cluttered background and irrelevant semantics, resulting in prominent feature contamination issue. Second, the recovered representations may remain incomplete and semantically ambiguous due to severely occluded scenarios.

\begin{figure*}[ttt]
  \centering
  %\color{blue}
  \includegraphics[width=1.0\textwidth]{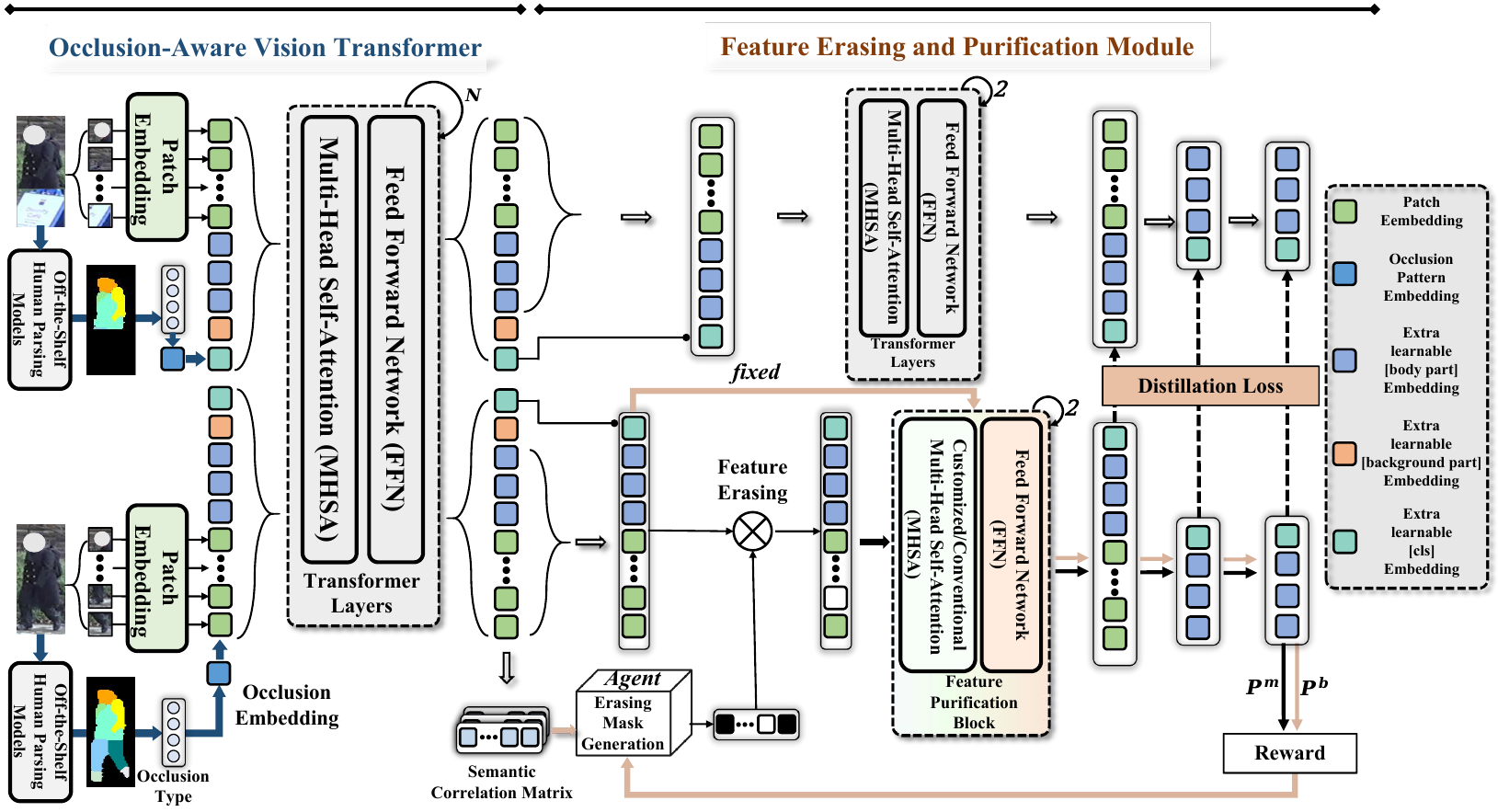}
  \caption{The overview of Occlusion-Guided Feature Purification Learning via Reinforced Knowledge Distillation approach. It contains a teacher branch for holistic person images and a student branch for occluded ones with OA-ViT and FEP modules.}
  \label{fig:arch}
\end{figure*}

Motivated by the above observations, we propose a novel Occlusion-Guided Feature Purification Learning via Reinforced Knowledge Distillation (OGFR), a teacher-student distillation framework that copes with diverse occlusion scenarios by learning occlusion-aware representation, and effectively excavates and transfers purified discriminative holistic knowledge from the holistic branch to the occluded branch through reinforced knowledge distillation for mitigating feature contamination. 
Specifically, as shown in Fig.~\ref{fig:arch}, OGFR consists of a teacher branch for holistic person images and a student branch for occluded ones, equipped with Occlusion-Aware Vision Transformer (OA-ViT) and Feature Erasing and Purification (FEP) module. 
As illustrated in Fig.~\ref{fig1}(b), OA-ViT facilitates precise and diverse occlusion type estimation under arbitrary occlusions, eliminating the constraints of predefined occlusion regions. This allows for more flexible occlusion augmentations, significantly improving the model’s generalization ability.
Additionally, OA-ViT encodes different occlusion types into learnable embeddings, incorporating explicit occlusion-related knowledge into the model. 
By enriching the input with both the original image and its corresponding occlusion conditions, OA-ViT transforms the model into an ``occlusion-aware" framework, allowing it to explicitly aware of the occlusion status for developing a deeper understanding of occlusion patterns and learning robust features even in complex occlusion scenarios.
Given that the features are incomplete when occlusion occurs, relying solely on occluded images is insufficient for comprehensive feature recovery. To address this, we introduce knowledge distillation, where occlusion-augmented data is input into the student branch to model occluded scenarios, while unaltered data is fed into the teacher branch to capture the holistic information. Through distillation, the student branch is guided to learn complementary features from the teacher branch, enhancing its ability to reconstruct identity-related representations.
Considering the inherently existing cluttered background and useless semantics in holistic images, we do not directly transfer the complementary knowledge from the holistic branch to the occluded branch, instead designing additional FEP module within the holistic branch. 
Unlike indiscriminately removing all tokens belonging to the background, FEP selectively eliminates noise and irrelevant information while retaining potentially valuable information.
This is achieved through a deep reinforcement learning-based erasing mask generation agent, which dynamically identifies and filters out low-quality patch tokens in holistic images, using the performance improvement over the case without the binary mask as the reward signal. 
By eliminating these low-quality tokens that negatively impact the model's performance, the agent improves feature purity and overall effectiveness.
To further refine representation learning, FEP replaces erased patch features with learnable embedding tokens, leveraging a customized two-layer Transformer decoder to extract potential information and purify discriminative cues.
After that, with the aid of knowledge distillation, the occluded branch assimilates such purified complementary knowledge to learn identity-related robust representation. Extensive experiments on the popular occluded and holistic ReID datasets demonstrate the effectiveness and superiority of OGFR.

The main contributions can be summarized as follows:
\begin{itemize}
    \item We propose a novel Occlusion-Guided Feature Purification Learning via Reinforced Knowledge Distillation (OGFR) approach for occluded person Re-ID. This teacher-student distillation framework effectively integrates diverse occlusion types into feature representation and facilitates the transfer of purified holistic knowledge from the holistic branch to the occluded branch.
    \item We introduce an Occlusion-Aware Vision Transformer that encodes visibility status of different body parts into learnable occlusion embeddings to guide occlusion-aware robust feature representation, ensuring adaptability to various occlusion scenarios.
    \item We design a Feature Erasing and Purification module within the holistic branch to identify low-quality patches through deep reinforcement learning and replace them with learnable embedding tokens for excavating refined identity-related information from holistic person images. 
    \item Our OGFR achieves state-of-the-art performance on several occluded and holistic ReID benchmarks, including Occluded-Duke \cite{miao2019pose}, Occluded-reID \cite{zhuo2018occluded}, P-DukeMTMC-reID \cite{zhuo2018occluded}, Market-1501 \cite{zheng2015scalable}, DukeMTMC-reID \cite{zheng2017unlabeled} and MSMT17 \cite{wei2018person}.
\end{itemize}

\section{Related Work}

\subsection{Holistic Person Re-Identification}
Over the past decade, a variety of holistic person Re-ID methods have sprung up, aiming at extracting discriminative and robust representations from full-body pedestrian images. Early studies \cite{zheng2011person,avraham2012learning,hirzer2012relaxed,chen2017beyond} primarily focused on designing hand-crafted features with body structures or robust distance metric learning. For example, Hirzer \textit{et al.} \cite{hirzer2012relaxed} employed metric learning to identify a suitable space for matching samples from different cameras. Instead of utilizing traditional Euclidean distance metrics, they adopted discriminative Mahalanobis distance measures, leading to a significant enhancement in performance. Recently, deep learning technique has been widely adopted for learning discriminative deep representations of holistic pedestrians. Sun \textit{et al.} \cite{sun2018beyond} proposed a part-based convolutional baseline for learning local features, which leverages a simple stripe partitioning strategy and refined part pooling to enhance intra-part consistency. Wang \textit{et al.} \cite{wang2018learning} proposed the Multiple Granularity Network (MGN), which consists of one branch for global feature representations and two branches for local feature representations, aiming at integrating discriminative information with various granularities. Zheng \textit{et al.} \cite{zheng2019pyramidal} developed a coarse-to-fine pyramid model via a multi-loss dynamic training scheme to incorporate local and global discriminative information. More recently, He \textit{et al.} \cite{he2021transreid} pioneered the use of pure transformer framework for object Re-ID, and proposed two novel modules to enhance the capacity of feature learning.

\subsection{Occluded Person Re-Identification}
Compared to holistic person Re-ID, occluded person Re-ID presents greater challenges. Existing approaches can be roughly divided into three categories: part-to-part matching methods, occlusion augmentation methods, and feature recovery methods. 
Part-to-part matching methods exploit predefined rules or external models to extract and align visible part representations from non-occluded regions to mitigate the occlusion effect.
For instance, Zhu \textit{et al.} \cite{zhu2020identity} clustered feature map pixels as pseudo-labels for body parts, enabling visible part-based representation learning.
However, these methods focus only on visible regions, completely ignoring the impact of different occlusion types.
To enhance robustness against various occlusion patterns, some studies \cite{wang2022feature} introduced occlusion augmentation, artificially generating occluded images to simulate real-world scenarios. Chen \textit{et al.} \cite{chen2021occlude} apply occlusions at four fixed positions and enforce occlusion-type score prediction to handle occlusions.
However, these techniques typically apply occlusions to fixed spatial regions and rely solely on raw visual content, forcing the model to infer occlusions implicitly, which restricts its adaptability to different complex occlusion scenarios.
Beyond occlusion augmentation, feature recovery methods seek to reconstruct missing information by leveraging visible near-neighbor features and holistic image representations.
Hou \textit{et al.} \cite{hou2021feature} proposed a Region Feature Completion block that captures long-range spatial and long-term temporal contexts for feature recovery.
Xu \textit{et al.} \cite{xu2022learning} designed a feature recovery transformer that leverages k-nearest neighbor features to restore occluded representations.
However, these approaches risk feature contamination, as holistic and near-neighbor features often include background noise and irrelevant semantics. 
Different from these existing methods, our proposed OGFR not only actively and adaptively handles various occlusion types but also simultaneously avoids feature contamination in holistic images.

\subsection{Knowledge Distillation}
Knowledge Distillation is a model compression technique that transfers knowledge from a high-capacity teacher model to a lightweight student model, enabling parameter reduction while maintaining performance. This technique is initially proposed by Hinton \textit{et al.} \cite{hinton2014distilling}, the fundamental idea is to guide the training of the student model using soft labels (probability distributions) generated by the teacher model, rather than relying solely on hard labels (single class labels). This enables the student model to acquire deep-level knowledge and enhance generalization capabilities from the teacher model, leading to improved performance.
Several studies \cite{zheng2021pose,hu2022deep,ni2023part,lan2023learning,jia2023semi} have applied knowledge distillation to person Re-ID. For example, Zheng \textit{et al.} \cite{zheng2021pose} utilized knowledge distillation to transfer knowledge from the pose-guided branch to the main branch during the training, which enables the discarding of pose-based branches while maintaining satisfied performance. In our approach, knowledge distillation is employed to address the challenges of occluded person Re-ID. We devised a teacher-student distillation framework where the teacher model operates on holistic images, while the student model specializes in handling occlusion scenarios. Through reinforced knowledge distillation, we effectively transfer purified, discriminative knowledge from the holistic branch to the occluded branch, enabling the student model to better handle occlusion scenarios.

\section{Proposed Method}

In this section, we provide a detailed introduction to our proposed method, Occlusion-Guided Feature Purification Learning via Reinforced Knowledge Distillation (OGFR). As depicted in Fig.~\ref{fig:arch}, OGFR comprises a dual-branch architecture, integrating a teacher branch and a student branch. The teacher branch integrates a weight-shared Occlusion-Aware Vision Transformer (OA-ViT) and a customized Feature Erasing and Purification (FEP) module, while the student branch consists of the same weight-shared Occlusion-Aware Vision Transformer along with two conventional transformer layers.

\subsection{Occlusion-Aware Vision Transformer}
In order to adaptively handle various occlusion scenarios, OA-ViT reformulates different occlusion types as the set of visibility status of body parts through an occlusion label estimator, and incorporates occlusion type clues into learnable occlusion embeddings to facilitate occlusion-aware robust feature learning via occlusion-aware vision encoder.

\subsubsection{Occlusion Label Estimator} 
\label{subsec:Occlusion Label Estimator}
Occlusion occurs when certain body parts of a pedestrian are obstructed by external objects, leading to partial loss of body integrity. Based on this, different occlusion types in pedestrian images can be uniformly represented as a set of visibility statuses of various body parts. 
Given a person image, we first generate parsing labels $\boldsymbol M \in \mathbb{R}^{H \times W \times (K+1)}$ with the pre-trained MASK-RCNN \cite{he2017mask} and PifPaf \cite{kreiss2019pifpaf} models, where $H,W$ denote the height and width of the image, and $K+1$ is the defined $K$ semantic body parts and background (For instance, $K=8$ defines the 
following semantic body parts: \textit{head}, \textit{left/right arm}, \textit{torso}, \textit{left/right leg} and \textit{left/right feet}). The element $\boldsymbol{M}[h,w,k] \in \{0, 1\}$ is the parsing result at spatial position $[h,w]$, indicating whether belonging to the $k$-th body part or to the background. Once we obtain the parsing labels $\boldsymbol{M}$ from the pre-trained model, we assign occlusion types based on the number of pixels within each body part as indicated by these labels.
However, due to domain discrepancies, the segmentation results from the parsing model are not always precise. To mitigate this issue and obtain a more stable occlusion type representation, we further merge these $K$ fine-grained semantic body parts into $C$ coarse-grained body parts ($K > C$). The corresponding coarse semantic parsing labels $\widetilde{\boldsymbol M} \in \mathbb{R}^{H \times W \times C}$ (excluding the background) are generated by the merging operation. It can be observed that if a body part is occluded by an obstacle, the integrity of that part is compromised and fewer pixels in the image belong to that body part. Considering that, we calculate the total number of pixels corresponding to  each body part to determine its visibility status, \textit{i.e.}, \textit{Occluded} or \textit{Non-Occluded}. It is formulated as follows:
\begin{equation}
\begin{aligned}
  \boldsymbol{y}(c) &=\sum_{h=0}^{H-1} \sum_{w=0}^{W-1} \widetilde{\boldsymbol{M}}[h,w,c]
\end{aligned}
\end{equation}
\begin{equation}
\begin{aligned}
l_c&=\begin{cases}
  1 & \text{if } \boldsymbol y(c)<\lambda,  \\
  0 & \text{otherwise.} 
\end{cases}
\end{aligned}
\end{equation}
where $\boldsymbol{y}(c)$ denotes the total number of pixels belong to $c$-th coarse-grained body part, $\lambda$ is a pre-defined threshold, and $l_c$ denotes the visibility status of $c$-th coarse-grained body part ($1$ for \textit{Occluded} and $0$ for \textit{Non-Occluded}). We finally acquire the occlusion type representation for arbitrary images, $\boldsymbol{z} = [l_0, l_1, ..., l_{C-1}] \in \mathbb{R}^{C}$. 

\subsubsection{Occlusion-Aware Vision Encoder}
We incorporate the obtained occlusion type information of the image into occlusion pattern embeddings, and insert them into Vision Transformer (ViT) \cite{dosovitskiy2021image} together with different types of embeddings, endowing the model with the capacity to perceive and handle diverse occlusion scenarios. Specifically, we initialize the learnable occlusion pattern embeddings as $\boldsymbol O_e\in \mathbb{R}^{N_0\times...\times N_{C-1}\times D}$, where $\{N_0, ..., N_{C-1}\}$ denotes the 2-dimension visibility status for each body part. Therefore, the occlusion pattern embeddings $\boldsymbol E_o\in \mathbb{R}^D$ for occlusion type $\boldsymbol z$ can be formulated as below:
\begin{equation}
    \boldsymbol E_o= \boldsymbol O_e\left [\boldsymbol z[0], ..., \boldsymbol z[C-1]\right ]
\end{equation}
In addition to the occlusion pattern embeddings, other essential types of embeddings, such as position embeddings and camera embeddings, are also appended to the input sequence. These provide crucial location and camera-specific information for the patch tokens in the sequence. As for patch tokens, we first split the image with a resolution of $H\times W$ into $N$ patches based on the following formulation:
\begin{equation}
    N=\left \lfloor \frac{H+S-P}{S}  \right \rfloor \times \left \lfloor \frac{W+S-P}S{}  \right \rfloor
\end{equation}
where $S$ and $P$ denote stride and patch size, $\left \lfloor  \cdot  \right \rfloor$ represents the floor function. Upon obtaining patch tokens $\boldsymbol P_s \in \mathbb{R}^{N\times D}$ through the projection operation, we introduce a learnable [cls] token $x_{cls}$ to capture the global representation. Moreover, $K+1$ learnable part tokens $\boldsymbol P_t \in \mathbb{R}^{(K+1)\times D}$ are appended to the input sequence for the adaptive learning of part representations, which are semantically aligned with $K$ fine-grained body parts and the background. Notably, we maintain $K$ fine-grained body parts segmentation instead of $C$ coarse-grained body parts segmentation to enhance the discriminative capabilities of these part representations. Finally, the input sequence is described as follows:
\begin{equation}
    \boldsymbol E=\left [\boldsymbol x_{cls};\boldsymbol P_t;\boldsymbol P_s \right ] + \boldsymbol P_E+\gamma_1  \boldsymbol E_o+\gamma_2 \boldsymbol C_e
\end{equation}
where $\boldsymbol x_{cls}$, $\boldsymbol P_t$ and $\boldsymbol P_s$ are concatenated to form the main body of the sequence. Note that the position of the learnable tokens $\boldsymbol P_t$ can be changed, \textit{e.g.}, by placing them right after the patch tokens or shuffling their internal order. For simplicity, we adopt the case where they are placed on the left.
%We simply represent it as the former case for simplification. 
Additionally, $\boldsymbol P_E$, $\boldsymbol E_o$, and $\boldsymbol C_e$ are directly added to these tokens. 
Here, $\boldsymbol P_E$ denotes position embeddings, $\boldsymbol C_e$ represents camera embeddings proposed by \cite{he2021transreid}. $\gamma_1$ and $\gamma_2$ are hyper-parameters used to balance the contributions of these embedding components. We feed the input sequence into the occlusion-aware Vision Transformer to obtain the robust representation $\boldsymbol F\in \mathbb{R}^{{(N+K+2)}\times D}$, consisting of \{$\boldsymbol f^g$,$\boldsymbol f^0$,$\boldsymbol f^1$,...,$\boldsymbol f^K$,$\boldsymbol f^p$\}, where $\boldsymbol f^g$ denotes the global feature, $\boldsymbol f^p\in \mathbb{R}^{N\times D}$ denotes the patch features, $\boldsymbol f^0$ denotes the part feature corresponding to the background, and \{$\boldsymbol f^1$,...,$\boldsymbol f^K$\} represent the part features corresponding to different body parts.

Based on the occlusion-aware ViT, we establish a teacher-student architecture, where the branch for holistic person images serves as the teacher and the branch for occluded ones acts as the student. Within this architecture, we extract $\boldsymbol F_h$ and $\boldsymbol F_o$ through the weight-shared occlusion-aware ViT, where $h$, $o$ denote the holistic and occluded branches, respectively. The holistic feature $\boldsymbol F_h$ is then refined via a Feature Erasing and Purification module to produce the purified feature $\boldsymbol F_{hp}$. In parallel, the occluded feature $\boldsymbol F_o$ is processed through two conventional transformer layers, resulting in the enhanced feature $\boldsymbol F_{op}$. These features, $\boldsymbol F_{hp}$ and $\boldsymbol F_{op}$, form the foundation for the subsequent 
 knowledge distillation process.

\subsection{Feature Erasing and Purification Module}
\begin{figure}[ttt]
   %\color{blue}
  \centering
  \includegraphics[width=0.47\textwidth]{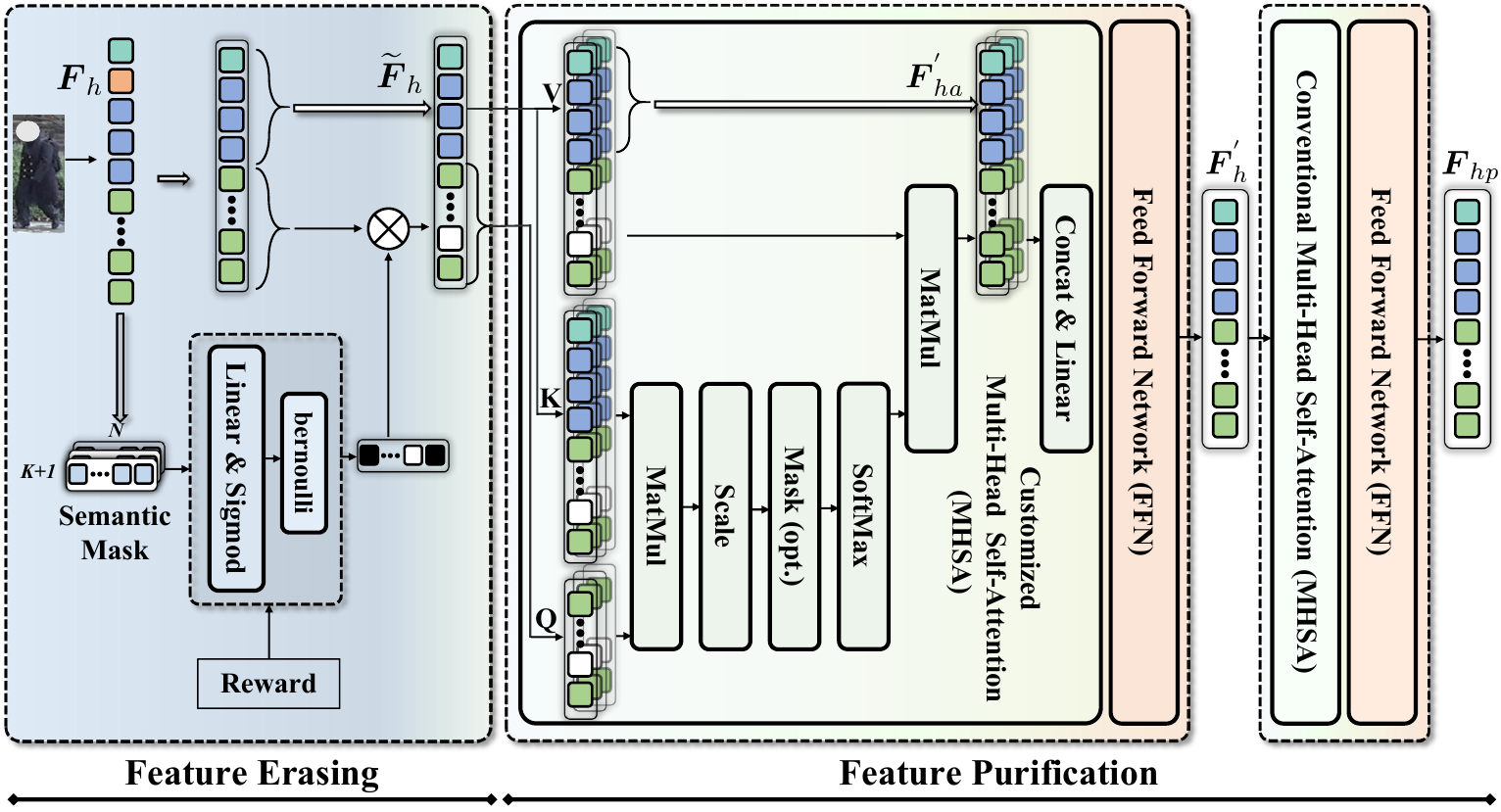}
  \caption{{The detailed structure of the Feature Erasing and Purification Module.}}
  \label{fig:arch_2}
\end{figure}

To explore meaningful discriminated clues while mitigating interference from irrelevant noise, we design the FEP Module within the holistic branch to excavate refined identity-related information in holistic images. As illustrated in Fig.~\ref{fig:arch_2}, FEP module trains an erasing mask generation agent that identifies low-quality patch tokens from all patch tokens and then substitutes these patch features with learnable embeddings to obtain purified discriminative representations via a customized two-layer decoder. 

\subsubsection{Erasing Mask Generation}
Given that the Markov Decision Process (MDP) provides a formal framework for modeling decision-making by defining key components essential for reinforcement learning, we formalize the process of erasing mask generation as a one-step MDP. Reinforcement learning is employed to enable the erasing mask generation agent to adaptively filter low-quality patch tokens based on the performance gains relative to the case without using the erasing mask.
The MDP is composed of state, agent, action, and reward, and is described as follows: 
\textit{State}: The semantic correlation matrix $\boldsymbol I\in \mathbb{R}^{N\times (K+1)}$, which encodes the affiliation relationship between different image patches and a specific body part, serves as the state representation. It is calculated by evaluating the similarity between patch features $\boldsymbol f_h^p$ and part features $ \{\boldsymbol f_h^k\}_{k=0}^K$.
\textit{Agent}: The agent generates an erasing mask using a fully connected (FC) layer followed by a sigmoid activation function, in which each element indicates the probability that a specific patch token is retained. It can be described as follows:
\begin{equation}
    \boldsymbol m=\text{sigmoid}(\boldsymbol I\boldsymbol G^T)
\end{equation}
where $\boldsymbol G\in \mathbb{R}^{1\times (K+1)}$ represents linear transformation.
\textit{Action}: The action determines whether each patch is selected, with values of 1 (to \textit{retain}) or 0 (to \textit{replace}). The decision follows a Bernoulli distribution, introducing stochasticity to prevent the agent from converging to suboptimal local solutions: \begin{equation}
   \boldsymbol a=\text{Bernoulli}(\boldsymbol m)
\end{equation}
\textit{Reward}: We devise a reward mechanism that provides incentives for using the erasing mask whose predicted scores for the true identity labels exceed those of the baseline. We follow the steps below to obtain baseline predictions. Without using the erasing mask, we directly feed the representation \{$\boldsymbol f_h^g$,$\boldsymbol f_h^1$,...,$\boldsymbol f_h^K$,$\boldsymbol f_h^p$\} into the customized two-layer decoder. Then the classifier outputs Re-ID prediction scores. We compute the average accuracy of predicted scores $p^b$ as the baseline. During the training of the agent, we keep the parameters of the baseline fixed and update its value every several epochs. Similarly, following the procedure above, we represent the predicted scores with using the erasing mask as $p^m$. The reward is defined as follows:
\begin{equation}
   r=\frac{p^m-p^b}{1-p^b}
\end{equation}
At each time step $t$ within an episode, the agent observes the state $s_t$, takes an action $a_t$ to generate the erasing mask $p^m$, and receives an immediate reward $r(s_t,a_t)$. The total reward for the $b$-th episode is given by $R_b=\sum_t r(s_t,a_t)$.
We employ the policy gradient method \cite{peters2010policy} to find the optimal set of parameters $\theta ^\ast$ representing the policy function, aiming to maximize the expected cumulative reward $\pi_{\theta}$. 
We define the optimization objective as $J(\pi_\theta )$. To approximate the gradient, we execute the agent for $B$ episodes and subsequently compute the average gradient, which can be described as follows:
\begin{equation}
    \bigtriangledown_{\theta} J(\theta )\approx \frac{1}{B} \sum_{b=1}^{B} \left ( \sum_{t=1}^{N} \bigtriangledown _\theta  \text{log}\pi_\theta(a_{b,t}\mid s_{b,t}) \right )R_b
\end{equation}
where $a_{b,t}$ is the action taken by the agent for the erasing mask in $b$-th episode.

\subsubsection{Feature Purification}
We determine whether to retain or replace the patch tokens based on the action $\boldsymbol a$. When the action is $0$, the corresponding low-quality patch tokens (patch features) are replaced with learnable embedding tokens. This replacement mechanism effectively discards redundant semantics and suppresses noise. After obtaining the substituted patch features with the remaining unchanged ones, we concatenate them with global and part features, which is denoted as $\widetilde{\boldsymbol F}_h$. This refined representation is subsequently fed into the customized two-layer decoder, formulated as follows:
\begin{equation}
    \begin{aligned}
       \boldsymbol F'_h&=\text{FFN}(\text{LN}(\text{MHSA}_{\text{fix}}(\widetilde{\boldsymbol F}_h)))\\
       \boldsymbol F_{hp}&=\text{FFN}(\text{LN}(\text{MHSA}(\boldsymbol F'_h)))
    \end{aligned}
\end{equation}
where $\text{FFN}$ and $\text{LN}$ denote a feed-forward layer and a layer normalization layer, respectively. $\text{MHSA}_{\text{fix}}$ is a specialized multi-head self-attention layer, which maintains the global and part features fixed and only allows the patch features learnable to thoroughly excavate potential identity-related clues. In $\text{MHSA}_{\text{fix}}$, each head independently processes the input features and derives the queries, keys, and values as follows:
\begin{equation}
        \boldsymbol Q=W_q\widetilde{\boldsymbol f}_h^p \quad
        \boldsymbol K=W_k\widetilde{\boldsymbol F}_h \quad
        \boldsymbol V=W_v\widetilde{\boldsymbol F}_h
\end{equation}
where $W_q$, $W_k$, $W_v$ are weights of linear projections, $\widetilde{\boldsymbol f}_h^p$ represents the patch features in $\widetilde{\boldsymbol F}_h$,  which are used to compute the queries  $\boldsymbol Q$. Through the attention mechanism, attention weights for $\boldsymbol Q$ relative to the keys $\boldsymbol K$ are computed, and these weights are utilized to aggregate $\boldsymbol V$, enabling the update of patch features:
\begin{equation}
    \begin{aligned}
        \boldsymbol F_{ha}^{'p}&=softmax(\frac{\boldsymbol Q \boldsymbol K^T}{\sqrt{k}}) \boldsymbol V
    \end{aligned}
\end{equation}
The fixed global and part features can be obtained by directly taking their corresponding values from $\boldsymbol V$, which can be represented by the following formula:
\begin{equation}
    \begin{aligned}
        \boldsymbol F_{ha}^{'fixed}&=\{\boldsymbol v^g,\boldsymbol v^1,...,\boldsymbol v^K\}
    \end{aligned}
\end{equation}
Subsequently, the updated patch features are concatenated with the fixed global and part features to construct the complete representation.
\begin{equation}
    \begin{aligned}
        \boldsymbol F_{ha}^{'} = [\boldsymbol F_{ha}^{'fixed}, \boldsymbol F_{ha}^{'p}] 
    \end{aligned}
\end{equation}
Following the above steps, each attention head produces an output feature $\boldsymbol{F}_{ha}^{'}$. The outputs from all heads are then concatenated and passed through a linear transformation to obtain the final representation of $\text{MHSA}_{\text{fix}}$. This is followed by \text{LN} and \text{FFN}, ultimately yielding the final refined representation $\boldsymbol{F}_{h}^{'}$.

Next, we apply \text{MHSA}, which denotes a conventional multi-head self-attention layer. This mechanism facilitates mutual learning among the global feature, the semantic part features, and the patch features, allowing them to interact and refine each other.
By integrating \text{MHSA}, \text{LN} and \text{FFN}, we obtain the refined purified discriminative feature $\boldsymbol F_{hp}$, which contains \{$\boldsymbol f_{hp}^g$,$\boldsymbol f_{hp}^1$,...,$\boldsymbol f_{hp}^K$,$\boldsymbol f_{hp}^p$\}.

\subsection{Knowledge Distillation}
To fully leverage identity-related clues embedded in the holistic branch, we employ knowledge distillation to transfer purified complementary knowledge from the holistic branch to the occluded branch. This process enables the occluded branch to assimilate holistic knowledge, thereby enhancing its capability to learn identity-related discriminative representation while effectively 
handling various occlusion scenarios. First of all, by feeding the features \{$\boldsymbol f_o^g$,$\boldsymbol f_o^1$,...,$\boldsymbol f_o^K$,$\boldsymbol f_o^p$\} into the two transformer layers, we obtain the features $\boldsymbol F_{op}$ for the occlusion branch, which is composed of \{$\boldsymbol f_{op}^g$,$\boldsymbol f_{op}^1$,...,$\boldsymbol f_{op}^K$,$\boldsymbol f_{op}^p$\}.
Then we employ the following loss functions to promote the transfer of knowledge.
\\(1) The mean square error loss is utilized to minimize the Euclidean distance between $\boldsymbol F_{hp}$ and $\boldsymbol F_{op}$ by:
\begin{equation}
    \mathcal{L}_{\text{mse}}=\left \| \boldsymbol F_{hp}-\boldsymbol F_{op} \right \|_2^2
\end{equation}
(2) We employ the cosine similarity loss to minimize the discrepancies in high-level global and part representations of two branches for enforcing feature consistency, which is formulated as:
\begin{equation}
    \mathcal{L}_{\text{cos}}=\frac{1}{K+1} \sum_{i}(1-\frac{\boldsymbol f_{hp}^i\cdot \boldsymbol f_{op}^i }{\left \| \boldsymbol f_{hp}^i \right \| \cdot \left \|  \boldsymbol f_{op}^i\right \| }), i\in\{g,1,...K\}
\end{equation}
(3) We incorporate the conventional knowledge distillation loss to further strengthen the knowledge transfer process, which is defined as follows:
\begin{equation}
    \mathcal{L}_{\text{kd}}=\frac{1}{K+1 } \sum_{i}(\alpha \mathcal{L}_{\text{id} }(\boldsymbol f_{op}^i)+\beta \text{KL} (\boldsymbol f_{hp}^i\parallel \boldsymbol f_{op}^i))
\end{equation}
where $i\in{g,1,...,K}$, and the identity loss $\mathcal{L}_{\text{id}}$ denotes cross-entropy loss that enforces alignment between the predicted logits, derived from the feature representation $\boldsymbol f_{op}^i$, and ground truth identity labels \cite{chen2021occlude}. \text{KL} denotes Kullback-Leibler Divergence.
$\alpha$ and $\beta$ are hyper-parameters to balance the loss terms.

\subsection{Model Optimization}
Firstly, we employ the identity loss to optimize the global features for excavating more discriminative fine-grained clues, which can be formulated as:
\begin{equation}
  \mathcal{L}_{\text{en}}=\mathcal{L}_{\text{id}}(\boldsymbol f_o^g)+\mathcal{L}_{\text{id}}(\boldsymbol f_h^g)
\end{equation}
In addition to the aforementioned loss, we impose supervision on the semantic correlation matrix using parsing labels $M$ in section~\ref{subsec:Occlusion Label Estimator} as the ground-truth. This is achieved via a cross-entropy loss, formulated as follows:
\begin{equation}
    \mathcal{L}_{\text{mask}}=-\sum_{k=0}^{K} \sum_{h=0}^{H-1} \sum_{w=0}^{W-1} {\boldsymbol M}_k(h,w)\text{log}(\widetilde {\boldsymbol I}_k(h,w))
\end{equation}
where $\widetilde {\boldsymbol I}\in \mathbb{R}^{H\times W\times (K+1)}$ denotes the transformed semantic correlation matrix obtained by performing rearrangement and bilinear interpolation operations on the matrix $\boldsymbol I$. $\widetilde {\boldsymbol I}_k(h,w)$ denotes the predicted probability for part $k$ at spatial location $(h,w)$. Moreover, we integrate the triplet loss \cite{schroff2015facenet} to encourage the holistic branch to learn more discriminative purified features, as formulated below:
\begin{equation}
\mathcal{L}_{\text{tr}}=\mathcal{L}_{\text{tri}}(\boldsymbol f_{hp}^g)+\frac{1}{K} \sum_{i=1}^{K} \mathcal{L}_{\text{tri}}(\boldsymbol f_{hp}^i)
\end{equation}
The overall objective function for optimization during the training stage is defined as follows:
\begin{equation}
\mathcal{L}=(\mathcal{L}_{\text{mse}}+\mu_1\mathcal{L}_{\text{cos}}+\mathcal{L}_{\text{kd}})+(\mathcal{L}_{\text{en}}+\mu_2\mathcal{L}_{\text{tr}})+\mathcal{L}_{\text{mask}}
\end{equation}
where $\mu_1$ and $\mu_2$ are hyper-parameters to adjust the corresponding loss terms.

During the inference stage, we employ the occluded branch to extract global and part features for calculating the similarity scores. 
Specifically, we only consider the similarity of visible features. The visibility label for the global feature is set to 1. 
For part features, visibility is assessed by examining pixel values in the transformed semantic correlation matrix $\widetilde {\boldsymbol I}$. A part is considered visible if the maximum pixel value within its corresponding region exceeds 0, formulated as follows:
\begin{equation}
\begin{aligned}
v_i&=\begin{cases}
  1 & \text{if } \max \limits_{h,w} \widetilde {\boldsymbol I_i}(h,w)> 0  \\
  0 & \text{otherwise.} 
\end{cases}
\end{aligned}
\end{equation}
where $i\in\{1,2,...K\}$. Inspired by \cite{miao2019pose}, the distance between visibility-based query images and gallery images is computed as follows:
\begin{equation}
    dist = \frac{ {\textstyle \sum_{i=1}^{K}(v_i^q\cdot v^g_i)\cdot D}(f_{op}^{i,q},f_{op}^{i,g})+D(f_{op}^{g,q}, f_{op}^{g,g})}{\textstyle \sum_{i=1}^{K}(v_i^q\cdot v^g_i)+1} 
\end{equation}
where $D(\cdot ,\cdot )$ refers to the Euclidean distance.

\section{Experiments}
In this section, we conduct extensive experiments on occluded and holistic datasets to verify the superiority of our proposed method. Then, we report ablation studies to analyze the effectiveness of each key component. Finally, we further provide the visualization results to intuitively show the effectiveness of our proposed method.

\subsection{Datasets and Evaluation Metrics}
For experimental evaluation, we conduct experiments on six public person ReID benchmarks, including three occluded datasets (Occluded-Duke\cite{miao2019pose}, Occluded-reID \cite{zhuo2018occluded}, P-DukeMTMC-reID \cite{zhuo2018occluded}) and three holistic datasets (Market-1501 \cite{zheng2015scalable}, DukeMTMC-reID \cite{zheng2017unlabeled} and MSMT17 \cite{wei2018person}). \textbf{Note:} The Duke-related datasets have been abandoned by proposers due to privacy concerns. In compliance with ethical and legal considerations, we commit to excluding any related dataset processing code and trained model parameters. The details of the datasets are summarized as follows.

\textbf{Occluded-Duke} \cite{miao2019pose} is an occluded Re-ID dataset containing 35,489 images from 8 cameras, with 15,618 for training and 2,210/17,661 for query/gallery.
\textbf{Occluded-reID} \cite{zhuo2018occluded} is a small-scale occluded dataset with 2,000 images of 200 identities, where each identity has five full-body and five occluded images.
\textbf{P-DukeMTMC-reID} \cite{zhuo2018occluded} is a occluded Re-ID dataset modified from DukeMTMC-reID \cite{zheng2017unlabeled}, comprising 24,143 images of 1,299 identities. The training set includes 12,927 images, while the testing set consists of 2,163 query images and 9,053 gallery images.
\textbf{Market-1501} \cite{zheng2015scalable} is a holistic Re-ID dataset with 12,936 training images and 3,368/15,913 for query/gallery, captured by 6 cameras.
\textbf{DukeMTMC-reID} \cite{zheng2017unlabeled} is a challenging holistic dataset containing 16,522 training images of 702 identities, with 2,228/17,661 for query/gallery.
\textbf{MSMT17} \cite{wei2018person} is a large-scale holistic dataset with 126,441 images of 4,101 identities from 15 cameras, including 32,621 for training and 11,659/82,161 for query/gallery.

\begin{table}[t]
%\color{blue}
\centering
\caption{Performance comparison with state-of-the-art methods on Occluded-Duke and Occluded-REID datasets. The first group: CNN-based methods; the second group: Transformer-based methods.
%in terms of Rank-1 (\%) and mAP (\%)
}
\label{table1}
\renewcommand\arraystretch{1.1}
\resizebox{\columnwidth}{!}{
\begin{tabular}{
>{\columncolor[HTML]{FFFFFF}}c |
>{\columncolor[HTML]{FFFFFF}}c |
>{\columncolor[HTML]{FFFFFF}}c 
>{\columncolor[HTML]{FFFFFF}}c |
>{\columncolor[HTML]{FFFFFF}}c 
>{\columncolor[HTML]{FFFFFF}}c }
\hline
{}    &  {}                              & \multicolumn{2}{c|}{{Occluded-Duke}} & \multicolumn{2}{c}{{Occluded-reID}} \\ \cline{3-6} 
\multirow{-2}{*}{{Method}} & \multirow{-2}{*}{{References}} & {Rank-1}                   & {mAP}                  & {Rank-1}                  & {mAP}                  \\ \hline
 PVPM \cite{gao2020pose} & CVPR 20                                            &  -                 &  -              &  70.4            &  61.2         \\
MHSA \cite{tan2022mhsa}   &     TNNLS 22                                           & 59.7             & 44.8           & -             &  -            \\
 Pirt \cite{ma2021pose}    &     ACM MM 21                                                    &  60.0                  &  50.9                 & -                   & -                   \\
IGOAS \cite{zhao2021incremental}   &     TIP 21                                                       &  60.1                  &  49.4                 &  81.1                 & -                 \\
RTGAT \cite{huang2023reasoning}    &     TIP 23                                                   &  61.0                  &  50.1                 &  71.8                 &  51.0                 \\ 
VGTri \cite{yang2021learning}    &     ICCV 21                                                   &  62.2                 &  46.3                 &  81.0                 &  71.0                 \\
OAMN \cite{chen2021occlude}    &     ICCV 21                                                &  62.6                  & 46.1               &  -                    &  -                    \\
 ISP \cite{zhu2020identity}    &     ECCV 20                                             &  62.8              &  52.3           &  -               &  -            \\
 PGFL-KD \cite{zheng2021pose}    &     ACM MM 21                                                   &  63.0                  &  54.1                 &  80.7                 &  70.3                 \\
 RFC \cite{hou2021feature}    &     TPAMI 21                                                       &  63.9                  &  54.5                 &  -                    &  -                    \\
 QPM \cite{wang2022quality}    &     TMM 23                                               &  64.4              &  49.7           &  -               &  -            \\
 OPR-DAAO \cite{wang2022occluded}    &     TIFS 22                                                       &  66.2                  &  55.4                 &  84.2                 &  75.1                \\
MoS \cite{jia2021matching}    &     AAAI 21                                                    & 66.6                  &  55.1                 &  -                   &  -                    \\ \hline
 PAT \cite{li2021diverse}    &     CVPR 21                                                       &  64.5                  &  53.6                 &  81.6                 &  72.1                 \\
 DRL-Net \cite{jia2022learning}    &     TMM 22                                                       &  65.8                  &  53.9                 &  -                 &  -                \\
 TransReID \cite{he2021transreid}    &     ICCV 21                                                 &  66.4                  &  59.2                 &  -                    &  -                    \\
 FED \cite{wang2022feature}    &     CVPR 22  & 68.1                 & 56.4    & 86.3   & 79.3                 \\
 AMG \cite{mao2023attention}    &     TMM 23   & 68.5      & 59.7                & -   &  -                 \\ 
SAP \cite{jia2023semi}    &     AAAI 23        & 70.0  & 62.2            & 83.0           &  76.8                 \\
FRT \cite{xu2022learning}   & TIP 22    & 70.7  & 61.3 & 80.4                 &  71.0                 \\ 
DPM \cite{tan2022dynamic}    &     ACM MM 22                                                       & 71.4                  & 61.8                 & 85.5                & 79.7                 \\
OAT \cite{li2024occlusion} & TIP 24    & 
 71.8 & 62.2 &82.6 & 78.2 \\
 MAHATMA \cite{zhang2025mask} & TCSVT 25 & 73.3 & 62.3 & 85.8  & 79.5  \\ \hline \hline
 \textbf{OGFR}       &                                                &  \textbf{76.6}          &  \textbf{64.7}        & \textbf{86.9}        &  \textbf{80.9}        \\
 \hline
\end{tabular}}
\end{table}

\begin{table*}[t]
%\color{blue}
\centering
\caption{Performance comparison with state-of-the-art methods on three holistic datasets including Market-1501, DukeMTMC-reID and MSMT17. The first group: CNN-based methods; the second group: Transformer-based methods. Hybrid denotes ResNet50 + transformer.
%in terms of Rank-1 (\%) and mAP (\%) 
}
\label{table2}
\renewcommand\arraystretch{1.1}
\resizebox{1.6\columnwidth}{!}{
\begin{tabular}{
>{\columncolor[HTML]{FFFFFF}}c |
>{\columncolor[HTML]{FFFFFF}}c |
>{\columncolor[HTML]{FFFFFF}}c |
>{\columncolor[HTML]{FFFFFF}}c 
>{\columncolor[HTML]{FFFFFF}}c |
>{\columncolor[HTML]{FFFFFF}}c 
>{\columncolor[HTML]{FFFFFF}}c |
>{\columncolor[HTML]{FFFFFF}}c 
>{\columncolor[HTML]{FFFFFF}}c }
\hline
{}                                  &  {}                                  & 
{}                                  & \multicolumn{2}{c|}{{Market-1501}} & \multicolumn{2}{c|}{{DukeMTMC-reID}} & \multicolumn{2}{c}{{MSMT17}} \\ \cline{4-9} 
\multirow{-2}{*}{{Method}} & \multirow{-2}{*}{{References}} & \multirow{-2}{*}{{Backbone}} & {Rank-1}       & {mAP}       & {Rank-1}        & {mAP}        & {Rank-1}    & {mAP}    \\ \hline
OAMN \cite{chen2021occlude}    &     ICCV 21    & ResNet50                                   & 93.2                  & 79.8               & 86.3                   & 72.6               & -                  & -               \\
IGOAS \cite{zhao2021incremental}    &     TIP 21     & ResNet50                            & 93.4                & 84.1              &  86.9                   &  75.1                &  -                  &  -               \\
 Pirt \cite{ma2021pose}    &     ACM MM 21   &  ResNet50-ibn                                  &  94.1                  &  86.3               &  88.9                   &  77.6                &  -                  &  -               \\
 MHSA \cite{tan2022mhsa}    &     TNNLS 22    &  ResNet50                                  &  94.6               &  84.0             &  87.3            &  73.1         &  -               &  -        \\
 OPR-DAAO \cite{wang2022occluded}    &     TIFS 22  &  ResNet50                                       &  95.1                  &  86.2               &  88.5                      &  76.5                   &  -                  &  -               \\
 RFC \cite{hou2021feature}    &     TPAMI 21      &  ResNet50                                  &  95.2                  &  89.2               &  90.7                   &  80.7                &  82.0               &  60.2            \\
 PGFL-KD \cite{zheng2021pose}    &     ACM MM 21  &  ResNet50                                &  95.3                  &  87.2               &  89.6                   &  79.5                &  -                  &  -               \\
 RTGAT \cite{huang2023reasoning}    &     TIP 23       &  ResNet50                                &  95.3                  &  88.2               &  89.1                   &  80.2                &  -                  &  -               \\
 ISP \cite{zhu2020identity}    &     ECCV 20    &  HRNet-W32                                   &  95.3               &  88.6             &  89.6            &  80.0         &  -               &  -            \\
 MoS \cite{jia2021matching}    &     AAAI 21      &  ResNet50-ibn                                  &  95.4                  &  89.0               &  90.6                   &  80.2                &  -                  &  -               \\ \hline
% MSDPA \cite{}    &     ACM MM 22      &  ViT-B                                  &  95.4                  &  89.5               &  90.9                   &  82.8                &  -                  &  -               \\
DRL-Net \cite{jia2022learning}    &     TMM 22      &  Hybrid                                   &  94.7                  &  86.9               &  88.1                   &  76.6                &  78.4                  &  55.3             \\
 FED \cite{wang2022feature}    &     CVPR 22       &  ViT-B                                 &  95.0                  &  86.3               &  89.4                   &  78.0                &  -                  &  -               \\
 AMG \cite{mao2023attention}    &     TMM 23        &  ViT-B                                 &  95.0                  &  88.5               &  -                      &  -                   &  -                  &  -               \\
 MAHATMA \cite{zhang2025mask} & TCSVT 25 & ViT-B & 95.2 &  88.2 & 91.2 & 81.2 & 85.6 &  \textbf{68.1} \\
 TransReID \cite{he2021transreid}    &     ICCV 21  &  ViT-B                                &  95.2                  &  88.9               &  90.7                   &  82.0                &  83.5               &  64.4            \\
 PAT \cite{li2021diverse}    &     CVPR 21      &  Hybrid                                  &  95.4                  &  88.0               &  88.8                   &  78.2                &  -                  &  -               \\
 FRT \cite{xu2022learning}    &     TIP 22       &  Hybrid                                  &  95.5                  &  88.1               &  90.5                   &  81.7                &  -                  &  -               \\
 DPM \cite{tan2022dynamic}    &     ACM MM 22    &  ViT-B                                  &  95.5                  &  89.7               &  91.0                   &  82.6                &  -                  &  -               \\
OAT \cite{li2024occlusion} & TIP 24    & ViT-B &
 95.7 & 89.9 &91.2 & 82.3 & - &-\\
 SAP \cite{jia2023semi}    &     AAAI 23      &  ViT-B                                  &  96.0                  &  \textbf{90.5}               &  -                      &  -                   &  85.7               &  67.8           \\ \hline \hline
 \textbf{OGFR}    &    &  ViT-B                                       &  \textbf{96.4}         & 90.2               &  \textbf{92.3}          &  \textbf{82.7}       & \textbf{86.2}                   &  66.9              \\ \hline
\end{tabular}}
\end{table*}

\begin{table}[t]
%\color{blue}
\centering
\caption{Performance comparison with state-of-the-art methods on P-DukeMTMC-reID dataset. The first group: CNN-based methods; the second group: Transformer-based methods. 
* indicates that the results are reproduced from the official code.
%in terms of Rank-1, Rank-5, Rank-10 (\%) and mAP (\%) 
}
\label{tablep}
\renewcommand\arraystretch{1.1}
\resizebox{1.0\columnwidth}{!}{
\begin{tabular}{
>{\columncolor[HTML]{FFFFFF}}c |
>{\columncolor[HTML]{FFFFFF}}c |
>{\columncolor[HTML]{FFFFFF}}c 
>{\columncolor[HTML]{FFFFFF}}c 
>{\columncolor[HTML]{FFFFFF}}c 
>{\columncolor[HTML]{FFFFFF}}c }
\hline
{}                                  & {}                                  & \multicolumn{4}{c}{{P-DukeMTMC-reID}}                                                                      \\ \cline{3-6} 
\multirow{-2}{*}{{Method}} & \multirow{-2}{*}{{References}} & {Rank-1} & {Rank-5} & {Rank-10} & {mAP}  \\ \hline
MHSA \cite{tan2022mhsa}         &           TNNLS 22                                           & 70.7           & 81.0            & 84.6             & 41.1          \\
PCB \cite{sun2018beyond}         &           ECCV 18                                        & 79.4            & 87.1           &  90.0             &  63.9          \\
 PVPM \cite{gao2020pose}         &           CVPR 20                                            &  85.1            & 91.3            &  93.3             &  69.9          \\
RTGAT \cite{huang2023reasoning}         &           TIP 23                                            &  85.6           & 91.5           &  93.4             & 74.3         \\
PGFA \cite{miao2019pose}         &           ICCV 19                                       & 85.7           &  92.0            &  94.2            &  72.4          \\
 ISP \cite{zhu2020identity}         &           ECCV 20                                            &  89.0            &  94.1            & 95.3             & 74.7          \\
QPM \cite{wang2022quality}      &           TMM 23                                               &  90.7            &  94.4            &  95.9             &  75.3          \\ \hline
FED$^{*}$ \cite{wang2022feature} & CVPR 22 & 90.9 & 94.4 & 95.4 & 79.4  \\
DPM$^{*}$ \cite{tan2022dynamic} & ACM MM 22 & 91.5 & 95.1 & 96.0 & 82.5 \\ 
TransReID$^{*}$ \cite{he2021transreid} & ICCV 21 & 92.4 & 95.8 & 96.9 & 82.6 \\ 
\hline \hline
 \textbf{OGFR}        &                                      &  \textbf{93.2}   & \textbf{96.2}   &  \textbf{97.1}   &  \textbf{83.4}\\ \hline
\end{tabular}}
\end{table}

\begin{table}[t]
\centering
\caption{Ablation studies of each component in OGFR on Occluded-Duke dataset. The OA-ViT refers to the Occlusion-Aware Vision Transformer, the FEP refers to the Feature Erasing and Purification Module, -S indicates the Student branch, -T the Teacher branch, the KD refers to the Knowledge Distillation.}
\label{table4}
\renewcommand\arraystretch{1.1}
\resizebox{\columnwidth}{!}{
\begin{tabular}{
>{\columncolor[HTML]{FFFFFF}}c 
>{\columncolor[HTML]{FFFFFF}}c 
>{\columncolor[HTML]{FFFFFF}}c 
>{\columncolor[HTML]{FFFFFF}}c |
>{\columncolor[HTML]{FFFFFF}}c 
>{\columncolor[HTML]{FFFFFF}}c 
>{\columncolor[HTML]{FFFFFF}}c 
>{\columncolor[HTML]{FFFFFF}}c }
\hline
\multicolumn{4}{c|}{{Method}}                                                                    & \multicolumn{4}{c}{{Occluded-Duke}}                                                                 \\ \cline{1-8}
{OA-ViT}                        & {FEP-S}       & {FEP-T}                 & {KD}                        & {Rank-1}           & {Rank-5}           & {Rank-10}          & {mAP}           \\ \hline
{-}                         & {-}         & {-}                & {-}                         & {65.8}          & {81.6}          & {86.5}          & {58.4}          \\
{\checkmark} & {-}            & {-}             & {-}                         & {68.5}          & {82.0}          & {87.0}          & {60.3}          \\
{\checkmark} & {\checkmark} & {-}         & {-}                & {72.5}          & {83.3}          & {86.8}          & {61.1}          \\
{\checkmark} & {\checkmark} & {-}         & {\checkmark}              & {73.1}          & {83.7}          & {86.8}          & {62.3}      \\
{\checkmark} & {\checkmark} & {\checkmark}         & {\checkmark}        & {74.3}          & {84.8}          & {87.3}          & {63.6}             \\
{\checkmark} & {-} & {\checkmark}  & {\checkmark} & {\textbf{76.6}} & {\textbf{86.4}} & {\textbf{89.8}} & {\textbf{64.7}} \\ \hline
\end{tabular}}
\end{table}

\textbf{Evaluation Metrics.} We adopt widely used Cumulative Matching Characteristic (CMC) at Rank-$k$ and mean Average Precision (mAP) as evaluation metrics. 

\subsection{Implementation Details}
We adopt ViT \cite{dosovitskiy2021image} model pre-trained on ImageNet \cite{deng2009imagenet} as our backbone network. Concretely, all input images are resized to $256 \times 128$. During training, we apply data augmentation techniques, including random horizontal flipping, padding pixels, random cropping and occlusion augmentation. For occlusion augmentation, we adopt the technique of selectively superimposing an obstacle patch onto a randomly chosen region within a pedestrian image, following previous methods \cite{jia2022learning,wang2022occluded}. We utilize SGD as the optimizer with the weight decay of $\text{1e-4}$. The learning rate is initiated as 0.008 with cosine learning rate decay, which is decayed at 40 and 70 epochs. The batch size is set to 64 with 4 images per identity. Following TransReID \cite{he2021transreid}, we adopt a smaller stride and set $\gamma_1$, $\gamma_2$ to 3.0. The dimension of the features $D$ is set to 768. The number of $K$ is set to 8. The number of $C$ is 4 containing the coarse-grained semantic regions \{head, arms, torso, legs\}. The hyper-parameter $\lambda$ is set to 5. The $\alpha$ and $\beta$ are set to 0.3 and 0.4. The $\mu_1$ and $\mu_2$ are set to 0.5. We implement our framework under two Nvidia Tesla v100 GPUs.

\subsection{Comparison with State-of-the-art Methods}

\subsubsection{Results on Occluded Datasets} 
To validate the effectiveness of our proposed OGFR, we conduct a comprehensive comparison with state-of-the-art methods across three occluded Re-ID datasets: Occluded-Duke, Occluded-REID, and P-DukeMTMC-reID. As shown in Table~\ref{table1} and Table~\ref{tablep}, the compared methods are categorized based on their architectural design into CNN-based and Transformer-based approaches.
Since there are no publicly available results for Transformer-based methods on the P-DukeMTMC-reID dataset, we reproduce several open-source methods to ensure a fair comparison. 
The CNN-based methods include PVPM \cite{gao2020pose}, MHSA \cite{tan2022mhsa}, Pirt \cite{ma2021pose}, IGOAS \cite{zhao2021incremental}, RTGAT \cite{huang2023reasoning}, VGTri \cite{yang2021learning}, OAMN \cite{chen2021occlude}, ISP \cite{zhu2020identity}, PGFL-KD \cite{zheng2021pose}, RFC \cite{hou2021feature}, QPM \cite{wang2022quality}, OPR-DAAO \cite{wang2022occluded}, MoS \cite{jia2021matching}, PCB \cite{sun2018beyond}, and PGFA \cite{miao2019pose}. Meanwhile, the Transformer-based methods encompass PAT \cite{li2021diverse}, DRL-Net \cite{jia2022learning}, TransReID \cite{he2021transreid}, FED \cite{wang2022feature}, AMG \cite{mao2023attention}, SAP \cite{jia2023semi}, FRT \cite{xu2022learning}, DPM \cite{tan2022dynamic}, OAT \cite{li2024occlusion} and MAHATMA \cite{zhang2025mask}. It is evident that Transformer-based approaches consistently outperform CNN-based methods, benefiting from their superior ability to model long-range dependencies and capture complex patterns. Our OGFR follows this trend, achieving state-of-the-art performance across all datasets.
Compared to the limited occlusion data-augmented method FED, OGFR achieves +8.5\% Rank-1 accuracy and +8.3\% mAP on Occluded-Duke, and +0.6\% Rank-1 accuracy and +1.6\% mAP on Occluded-REID. It also obtains an improvement of +2.3\% Rank-1 accuracy and +4.0\% mAP on the P-DukeMTMC-reID dataset, demonstrating its superior performance. The substantial gain is attributed to OA-ViT, which incorporates the visibility status of body parts into learnable occlusion embeddings, effectively guiding occlusion-aware feature learning for robust identity representation across varying occlusion scenarios.
Additionally, compared to SAP, which also adopts a teacher-student architecture with ViT as the backbone, OGFR achieves +6.6\% Rank-1 accuracy and +2.5\% mAP on Occluded-Duke, and +3.9\% Rank-1 accuracy and +4.1\% mAP on Occluded-REID, further validating its superiority. This demonstrates the effectiveness of refined holistic knowledge transferred from the teacher branch. Simultaneously, the student branch efficiently absorbs purified knowledge to learn identity-related discriminative features, thereby achieving outstanding performance.

\subsubsection{Results on Holistic Datasets} To further verify the superiority of OGFR, we conduct extensive experiments on three holistic datasets: Market-1501, DukeMTMC-reID, and MSMT17. We compare OGFR against state-of-the-art methods, which are categorized into CNN-based and Transformer-based methods, including OAMN \cite{chen2021occlude}, IGOAS \cite{zhao2021incremental}, Pirt \cite{ma2021pose}, MHSA \cite{tan2022mhsa}, RFC \cite{hou2021feature}, PGFL-KD \cite{zheng2021pose}, RTGAT \cite{huang2023reasoning}, ISP \cite{zhu2020identity}, MoS \cite{jia2021matching}, DRL-Net \cite{jia2022learning}, FED \cite{wang2022feature}, AMG \cite{mao2023attention}, MAHATMA \cite{zhang2025mask}, OPR-DAAO \cite{wang2022occluded}, TransReID \cite{he2021transreid}, PAT \cite{li2021diverse}, FRT \cite{xu2022learning}, DPM \cite{tan2022dynamic}, OAT \cite{li2024occlusion}, SAP \cite{jia2023semi}. As depicted in Table~\ref{table2}, our proposed OGFR achieves the best Rank-1 accuracy of 96.4\% and 86.2\% on the Market-1501 and MSMT17 datasets, respectively. It also achieves near-optimal mAP on these datasets. Compared to the latest MAHATMA method, OGFR surpasses it by +2.0\% mAP on Market-1501, demonstrating its strong capability in holistic person retrieval. Although it falls slightly behind by 1.2\% mAP on the MSMT17 dataset, OGFR still maintains highly competitive performance, highlighting its robustness across diverse datasets. Furthermore, OGFR attains the highest Rank-1 accuracy of 92.3\% and mAP of 82.7\% on the DukeMTMC dataset. The success on these datasets demonstrates that our method not only handles occlusion effectively but also excels at capturing discriminative information in non-occluded scenarios.

\subsection{Ablation Studies}
\subsubsection{Effectiveness of Proposed Modules} 
We conducted ablation studies on Occluded-Duke dataset to validate the effectiveness of each component. As shown in Table~\ref{table4}, The baseline is a single-branch model with occlusion augmentation but without OA-ViT, FEP, or KD, using the ViT \cite{dosovitskiy2021image} framework with identity and triplet losses.
Incorporating OA-ViT improves the baseline by +2.7\% in Rank-1 accuracy and +1.9\% in mAP due to its ability to encode diverse occlusion patterns into occlusion pattern embeddings for robust feature learning. Adding the FEP module to the single-branch model further boosts performance by +4.0\% in Rank-1 accuracy and +0.8\% in mAP, as FEP adaptively filters low-quality patch tokens and replaces them with learnable embedding tokens to extract purified discriminative features.
Along with the introduction of knowledge distillation (KD), the model transitions into a teacher-student structure, where the complete branch serves as the teacher and the occlusion branch as the student.
Retaining FEP in the student branch improves Rank-1 accuracy by +0.6\% and mAP by +1.2\%, showing that the teacher provides valuable supplementary information. However, placing FEP in the teacher branch achieves better results.
Further analysis shows that placing FEP exclusively in the teacher branch is optimal. The teacher branch refines holistic knowledge to obtain purer knowledge, while the student branch, without FEP, retains complete occluded image information, enabling it to leverage the rich information to learn and absorb more complete and purer knowledge from the teacher.

\begin{figure}[t]
  \centering
  \includegraphics[width=0.85\columnwidth]{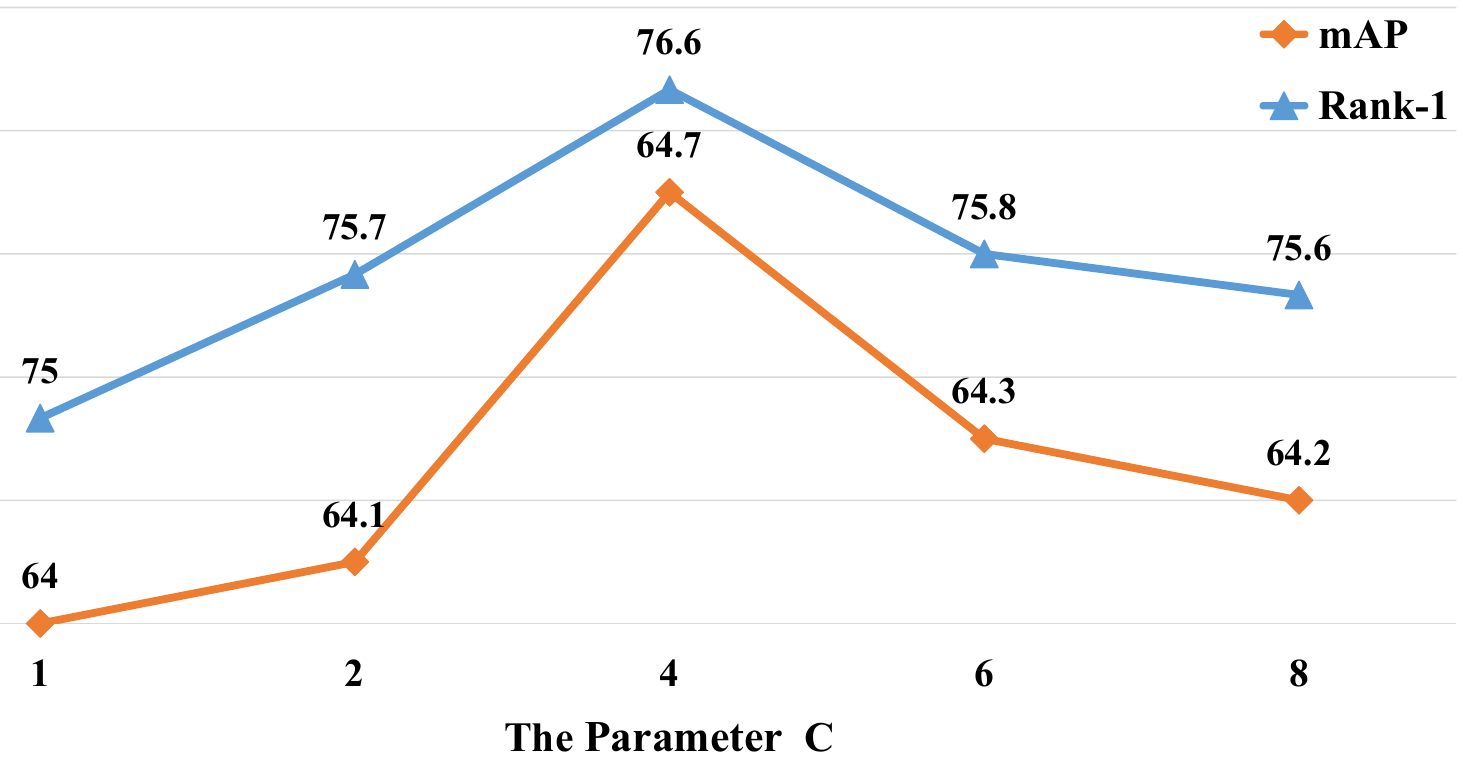}
  \caption{Impact of the parameter $C$ for the occlusion type in terms of Rank-1 (\%) and mAP (\%) on Occluded-Duke dataset.}
  \label{fig3}
  \end{figure}

\begin{table}[t]
%\color{blue}
\centering
\tiny
\caption{Performance comparison with different types of masks on Occluded-Duke dataset. Here, w/o denotes without Erasing Mask Generation, the \begin{math}\mathcal{R}\end{math} denotes Random Mask Generation, the \begin{math}\mathcal{A}\end{math} denotes Background and Occlusion Mask Generation, the \begin{math}\mathcal{E}\end{math} denotes our Erasing Mask Generation.}
\label{table5}
\renewcommand\arraystretch{1.1}
\resizebox{\columnwidth}{!}{
\begin{tabular}{
>{\columncolor[HTML]{FFFFFF}}c |
>{\columncolor[HTML]{FFFFFF}}c 
>{\columncolor[HTML]{FFFFFF}}c 
>{\columncolor[HTML]{FFFFFF}}c 
>{\columncolor[HTML]{FFFFFF}}c }
\hline
                                       & \multicolumn{4}{c}{Occluded-Duke} \\ \cline{2-5} 
\multirow{-2}{*}{Mask Method} & Rank-1             & Rank-5            & Rank-10           & mAP            \\ \hline
w/o                                                            & 74.5            & 86.3           & 89.5           & 64.3           \\
\begin{math}\mathcal{R}\end{math}                                                         & 74.4            & 85.8           & 88.9           & 63.7           \\
\begin{math}\mathcal{A}\end{math}                                                         & 75.4            & 86.2           & 89.6           & 64.3           \\
\begin{math}\mathcal{E}\end{math}                                                             & \textbf{76.6}   & \textbf{86.4}  & \textbf{89.8}  & \textbf{64.7}  \\ \hline
\end{tabular}}
\end{table}

\begin{table}[t]
\centering
\tiny
\caption{Performance comparison between different types of decoders on Occluded-Duke dataset. Here, the \begin{math}\mathcal{M}\end{math} denotes a conventional decoder with two multi-head self-attention layers, while \begin{math}\mathcal{S}\end{math} denotes our customized decoder.}
\label{table6}
\renewcommand\arraystretch{1.1}
\resizebox{0.95\columnwidth}{!}{
\begin{tabular}{
>{\columncolor[HTML]{FFFFFF}}c |
>{\columncolor[HTML]{FFFFFF}}c 
>{\columncolor[HTML]{FFFFFF}}c 
>{\columncolor[HTML]{FFFFFF}}c 
>{\columncolor[HTML]{FFFFFF}}c }
\hline
{}                                  & \multicolumn{4}{c}{{Occluded-Duke}}                                                                 \\ \cline{2-5} 
\multirow{-2}{*}{{Method}} & {Rank-1}        & {Rank-5}        & {Rank-10}       & {mAP}           \\ \hline
{\begin{math}\mathcal{M}\end{math} }                                                         & {75.3}          & {86.0}          & {89.0}          & {64.1}          \\
{\begin{math}\mathcal{S}\end{math}}                                                         & {\textbf{76.6}} & {\textbf{86.4}} & {\textbf{89.8}} & {\textbf{64.7}} \\ \hline
\end{tabular}}
\end{table}

\subsubsection{Analysis of the Occlusion Type in OA-ViT} We analyze the key parameter $C$ of the occlusion type in OA-ViT. As shown in Fig.~\ref{fig3}, when the parameter $C$ increases from 1 to 4, the model demonstrates an improvement in performance from 75.0\% to 76.6\% in Rank-1 accuracy. This indicates that using visibility status of finer granularity body parts allows for accurately characterize occlusion patterns, leading to superior retrieval performance. However, when increasing the parameter $C$ from 4 to 8, OGFR achieves slight performance degradation, signifying that excessive occlusion types introduce redundant information. Thus, we set the parameter $C$ to 4 in our experiments, which achieves optimal performance.

\begin{figure*}[t]
  \centering
  \includegraphics[width=2\columnwidth]{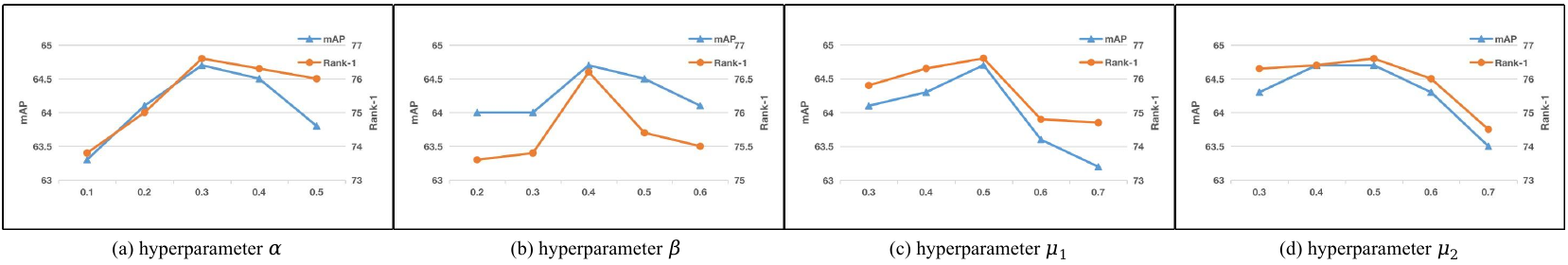}
  \caption{Performance curve of four hyper-parameters within the loss functions on Occluded-Duke dataset. (a) Performance curve of the hyperparameter $\alpha$, (b) Performance curve of hyperparameter $\beta$, (c) Performance curve of hyperparameter $\mu_{1}$, (d) Performance curve of hyperparameter $\mu_{2}$.}
  \label{fig6}
\end{figure*}

\subsubsection{Analysis of Erasing Mask Generation in FEP} In Table~\ref{table5}, we compare performance under different mask generation methods in the FEP module. It can be observed that the performance of random mask generation is reduced by +0.6\% mAP compared to without the mask. This suggests that random mask generation could potentially obscure crucial information. On the other hand, masking all background and occlusion tokens leads to a +0.9\% improvement in Rank-1 accuracy compared to without the mask, indicating that it helps reduce some noise interference. 
Our erasing mask generation improves Rank-1 accuracy by +2.2\% over random masking generation  and +1.2\% over background and occlusion masking generation.
By adaptively avoiding the interference of cluttered backgrounds and irrelevant semantics while retaining useful information, the erasing mask generation method proves to be more effective than random masking and background/occlusion masking, leading to superior performance.

\subsubsection{Analysis of Customized Decoder in FEP} As illustrated in Table~\ref{table6}, we conduct a comparative performance analysis across different types of decoders within the FEP module. It can be observed that our customized decoder exhibits a noteworthy improvement, achieving +1.3\% increase in Rank-1 accuracy and +0.6\% in mAP compared to the conventional decoder. The improvement can be attributed to the incorporation of a specialized multi-head self-attention layer. This layer maintains the global and part features as fixed while allowing the patch features to be learnable. Such a design facilitates patch features in thoroughly excavating potential identity-related clues from global and semantic features. More importantly, it prevents patch features from introducing noise or contaminating the global and semantic context.

\subsubsection{Analysis of Hyper-parameter within Loss Functions} To validate the robustness of hyper-parameters within the loss function, we examine the impact of varying hyperparameter values on the performance of OGFR, as illustrated in Fig. \ref{fig6}.

\textbf{Influence of Hyperparameter $\alpha$.}
We evaluate the effect of hyperparameter $\alpha$ on performance. The parameter $\alpha$ is the coefficient of the identity loss ($\mathcal{L}_{\text{id}}$) within the knowledge distillation loss ($\mathcal{L}_{\text{kd}}$). As depicted in Fig. \ref{fig6} (a), too small $\alpha$ results in poor performance. This is primarily because a lower $\alpha$ weakens the preservation of crucial identity-related information during knowledge distillation. Conversely, too large $\alpha$ may disrupt the delicate balance required for effective knowledge distillation, which degrades the performance. Therefore, in our experiments, $\alpha$ is set to 0.3.

\textbf{Influence of Hyperparameter $\beta$.}
 We investigate the influence of hyperparameter $\beta$, which controls the weight assigned to the Kullback-Leibler (KL) divergence term in the knowledge distillation loss. As shown in Fig. \ref{fig6} (b), small $\beta$ results in insufficient consideration of the knowledge distillation term, thereby impairing performance. While larger $\beta$ could potentially overshadow the impact of other loss components, causing instability and degrading the performance. Thus, in our experimental setup, $\beta$ is set to 0.4.

\textbf{Influence of Hyperparameter $\mu_1$.}
We analyze the impact of hyperparameter $\mu_1$ within the overall loss function, which controls the contribution of the cosine similarity loss ($\mathcal{L}_{\text{cos}}$). As illustrated in Fig. \ref{fig6} (c), small $\mu_1$ diminishes the significance of the cosine similarity loss, while a larger value can dominate the overall loss function, adversely affecting discriminative feature learning. Thus, we set $\mu_1$ to 0.5 in our experimental setup.

\begin{table}[t]
\centering
\caption{Ablation studies of each loss function in Knowledge Distillation on Occluded-Duke dataset.}
\label{table7}
\renewcommand\arraystretch{1.1}
\resizebox{\columnwidth}{!}{
\begin{tabular}{
>{\columncolor[HTML]{FFFFFF}}c 
>{\columncolor[HTML]{FFFFFF}}c 
>{\columncolor[HTML]{FFFFFF}}c |
>{\columncolor[HTML]{FFFFFF}}c 
>{\columncolor[HTML]{FFFFFF}}c 
>{\columncolor[HTML]{FFFFFF}}c 
>{\columncolor[HTML]{FFFFFF}}c }
\hline
\multicolumn{3}{c|}{{Knowledge Distillation}}       & \multicolumn{4}{c}{{Occluded-Duke}}                                                                 \\ \cline{1-7}
{MSE}   & {COS}   & {KD}    & {Rank-1}        & {Rank-5}        & {Rank-10}       & {mAP}           \\ \hline
{-}          & {\checkmark} & {\checkmark} & {74.2}          & {84.9}          & {88.5}          & {63.3}          \\
{\checkmark} & {-}          & {\checkmark} & {75.1}          & {86.0}          & {89.0}          & {63.6}          \\
{\checkmark} & {\checkmark} & {-}          & {74.5}          & {85.9}          & {89.4}          & {62.7}          \\
{\checkmark} & {\checkmark} & {\checkmark} & {\textbf{76.6}} & {\textbf{86.4}} & {\textbf{89.8}} & {\textbf{64.7}} \\ \hline
\end{tabular}}
\end{table}

\begin{table}
%\color{blue}
\centering
%\scriptsize
\caption{Complexity comparison with state-of-the-art methods on the occluded-duke dataset. Here, M represents Million, G represents Giga, and the units for training time ($T_T$) and inference time ($T_I$) are both expressed in ms/image.}
\label{table8}
\renewcommand\arraystretch{1.1}
\resizebox{\columnwidth}{!}{
\begin{tabular}{
>{}c |
>{}c 
>{}c 
>{}c 
>{}c
>{}c
>{}c}
\hline
{Method}          & \multicolumn{1}{c}{{Rank-1}} & \multicolumn{1}{c}{{mAP}} & \multicolumn{1}{c}{{Params}} & \multicolumn{1}{c}{{FLOPs}} & \multicolumn{1}{c}{{T$_{T}$}} & \multicolumn{1}{c}{{T$_{I}$}} \\ \hline
{TransReID \cite{he2021transreid}} & {66.4}   & {59.2}     & {103.3M}         & {23.6G} & 4.8 &   5.2 \\ 
{FED \cite{wang2022feature}} & {68.1}    & {56.4}    & {146.1M}  & {23.2G}    &  17.3 & 3.5 \\ 
{DPM \cite{tan2022dynamic}} & {71.4}  & {61.8}     & {146.1M}  & {32.8G} & 12.8 & 14.1 \\ \hline
{\textbf{OGFR}}      & {\textbf{76.6}} & {\textbf{64.7}}    & {121.7M}  & {26.5G}  & 14.1 &  7.0  \\ \hline
\end{tabular}}
\end{table}

\begin{figure}[t]
  %\color{blue}
  \centering
  \includegraphics[width=1.0\columnwidth]{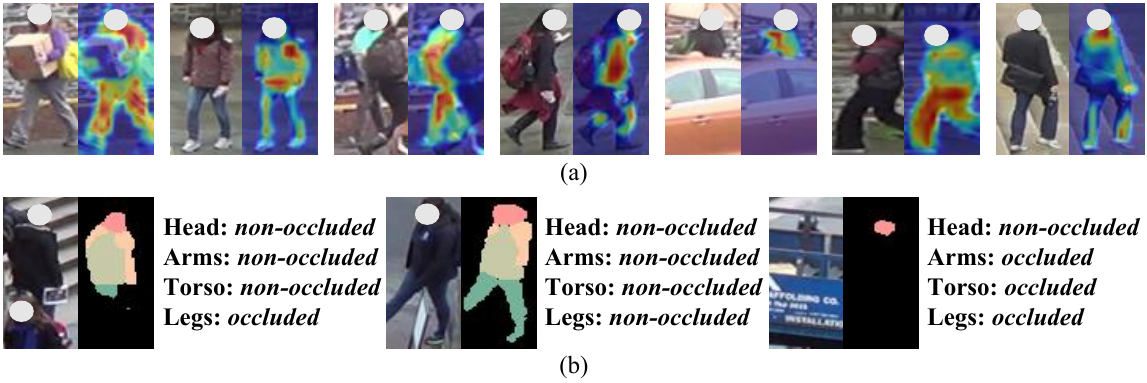}
  \caption{(a) Feature visualization results of the learned representation from the occluded branch. (b) Visualization of occlusion types for four coarse-grained body parts including \{\textit{head, arms, torso, legs}\}.}
  \label{fig4}
\end{figure}

\begin{figure*}
%\color{blue}
\centering
\includegraphics[width=0.75\textwidth]{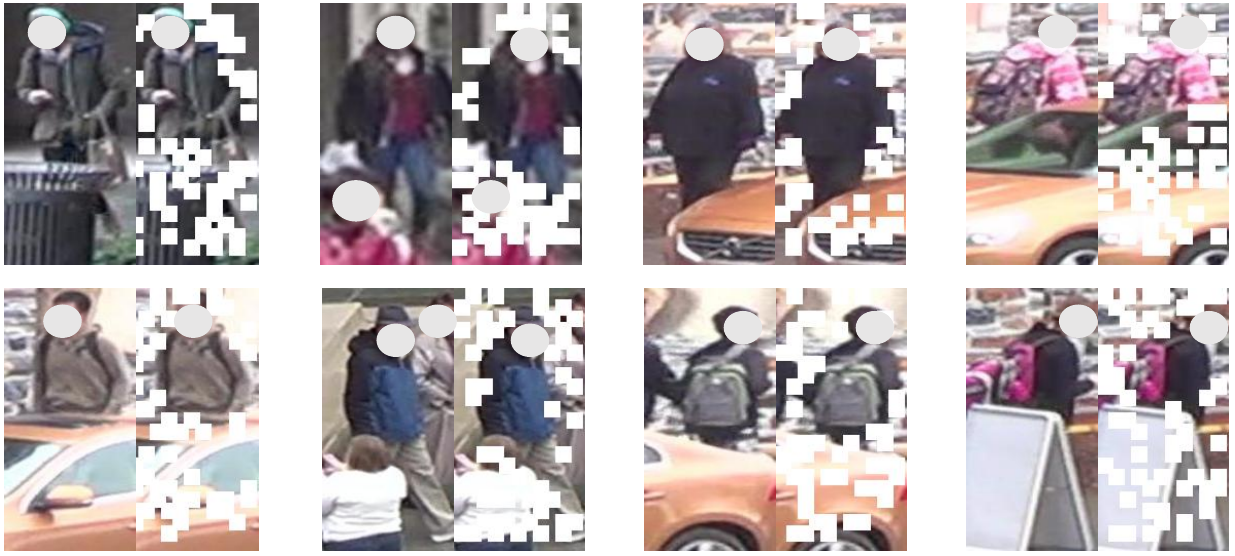}
\captionsetup{justification=centering}
\caption{The visualization results of erasing mask in FEP. The white regions represent the areas selected by the agent for masking.}
\label{fig:2}
\end{figure*}

\begin{figure*}[ttt]
    \centering
    %\color{blue}
    \includegraphics[width=0.75\textwidth]{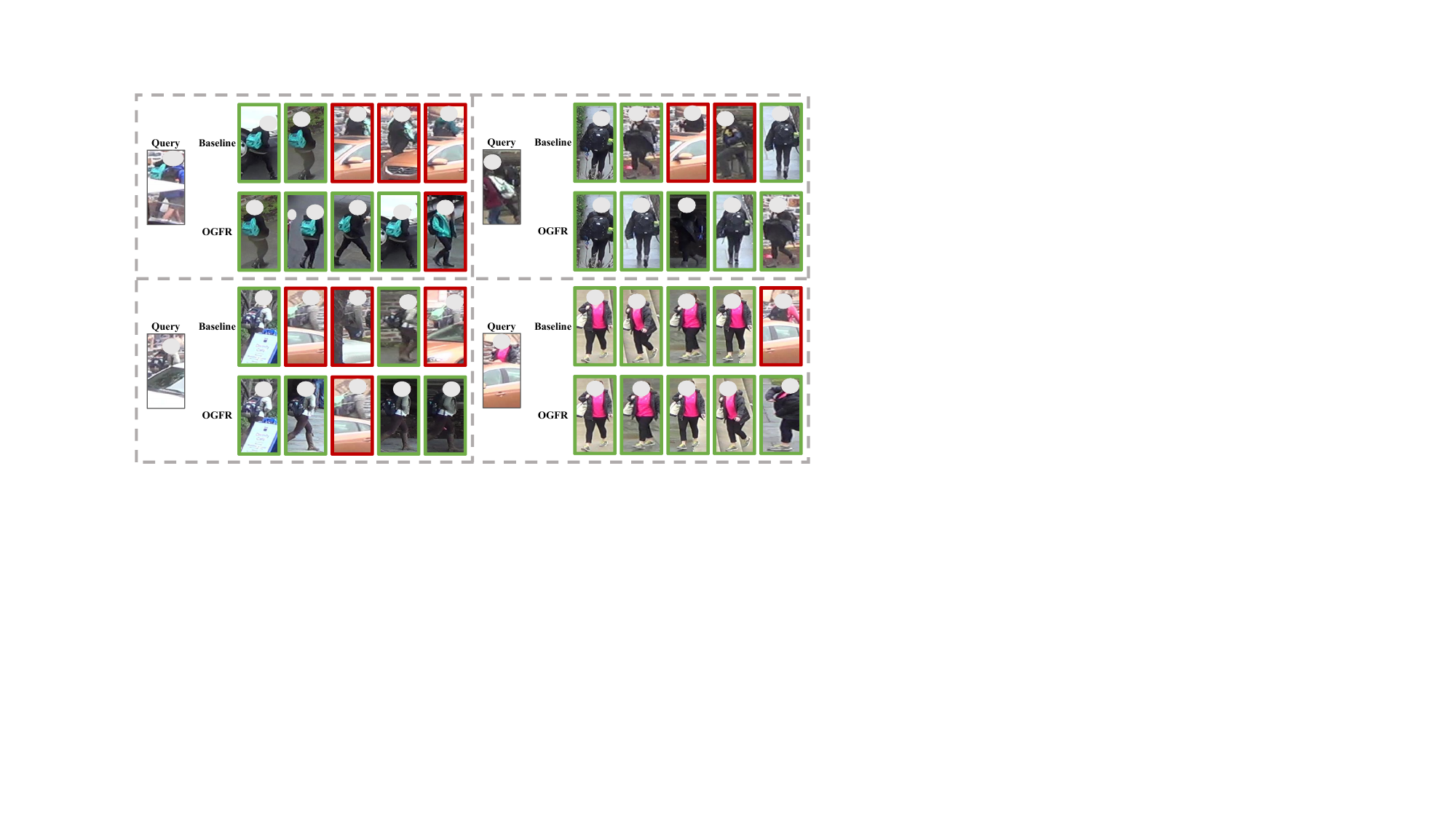}
    \caption{Comparison of Rank-5 retrieval results between the baseline and our proposed OGFR on the Occluded-Duke dataset. The green and red rectangles highlight positive and negative matching, respectively.}
    \label{fig:1}
\end{figure*}

\textbf{Influence of Hyperparameter $\mu_2$.}
We analyze the influence of hyperparameter $\mu_2$, which regulates the contribution of the triplet loss ($\mathcal{L}_{\text{tr}}$) within the overall loss function. As shown in Fig. \ref{fig6} (d), $\mu_2=0.5$ achieves the best performance. This is because small $\mu_2$ diminishes the influence of the triplet loss, potentially impairing the ability to learn discriminative features, while larger $\mu_2$ could overly emphasize the triplet loss, potentially causing an imbalance in the learning process.

Finally, it is worth noting that these four hyper-parameters exhibit satisfactory robustness. Across variations in their values, model performance consistently remains within a range of no more than 3\%. This observation indicates that, regardless of parameter adjustments, the model maintains a relatively consistent level of robustness in terms of performance stability.

\subsubsection{Analysis of Knowledge Distillation} As shown in Table~\ref{table7}, we conduct an ablation analysis on the three losses in knowledge distillation. The performance is improved by +2.4\% Rank-1 accuracy and +1.4\% mAP with the mean squared error (MSE) loss, as the MSE loss minimizes the feature discrepancy between the holistic branch and the occluded branch. This ensures that the features of both branches are as similar as possible, thereby aligning the feature spaces of the two branches. Furthermore, compared with the combination of the MSE loss and knowledge distillation (KD) loss, the performance is improved by +1.5\% Rank-1 accuracy and +1.1\% mAP using the cosine similarity (COS) loss, which reduces divergence between global and semantic part representations within the two branches, promoting orientation coherence in high-level features.
%This is because the COS loss minimizes the divergence between the global and semantic part representations within the two branches. This ensures the coherence of orientation in high-level feature representations. 
Finally, with the assistance of the KD loss, the performance is improved by +2.1\% Rank-1 accuracy and +2.0\% mAP. This is because the KD loss minimizes the difference between the probability distributions of the holistic branch and the occluded branch, encouraging the occluded branch to adhere to the distribution of the holistic branch.

\subsection{Visualization Results}
As shown in Fig.~\ref{fig4} (a), each pair of images represents the original image and the visualization of features from the occluded branch, respectively. We can observe OGFR is not affected by occlusions, allowing it to focus on more discriminative semantic clues. Moreover, we visualize the visibility status of four body parts under different occlusions in Fig.~\ref{fig4} (b). By integrating the visibility information of body parts into the learnable occlusion pattern embeddings, OGFR effectively handles various occlusion scenarios. 
Furthermore, Fig.~\ref{fig:2} presents the visualization results of the erasing mask, which effectively identifies obstacles, cluttered backgrounds, and other non-target pedestrians. 
This is because we employ an erasing mask generation agent to dynamically identify low-quality patch tokens from all patch tokens in a deep reinforcement learning manner based on the performance gains relative to the case without employing the erasing mask.

\subsection{Qualitative Retrieval Results}
To demonstrate the effectiveness of our OGFR in handling occluded scenes, we present the comparison of Rank-5 retrieval results between the baseline and our proposed method in Fig.~\ref{fig:1}, where green and red rectangles represent correct and incorrect retrieval results, respectively. For each occluded query person image on the left, where the occlusion scenarios have not appeared in the training data, our OGFR demonstrates a strong capability to accurately retrieve the target pedestrians despite the interference of various occlusion conditions, compared to the baseline. Furthermore, as observed from the first two retrieval examples in the top row, OGFR exhibits enhanced perceptual ability in identifying target pedestrians in cases of occlusion by other individuals, resulting in superior performance. This highlights the robustness and effectiveness of OGFR in addressing real-world occlusion scenarios within pedestrian retrieval tasks.

 \subsection{Computational Complexity}
As shown in Table~\ref{table8}, we conduct extensive experiments to demonstrate that our OGFR not only exhibits superior performance but also advantages in terms of model parameters, floating-point operations (FLOPs), training time (T$_T$), and inference time (T$_I$). 
We compare OGFR with recent popular occluded Re-ID methods: TransReID \cite{he2021transreid}, FED \cite{wang2022feature}, and DPM \cite{tan2022dynamic}, ensuring a fair comparison by using ViT-B as the backbone. 
All experiments are conducted on one NVIDIA GeForce GTX 1080Ti. From Table~\ref{table8}, it can be observed that OGFR exhibits relatively fewer model parameters while maintaining comparable FLOPs compared to the other methods. Compared to DPM \cite{tan2022dynamic}, which does not rely on external models, our approach showcases advantages in both model parameters and FLOPs. 
Furthermore, in terms of training times, our approach, which incorporates additional modules like reinforcement learning and knowledge distillation, has a longer training time compared to TransReID. However, it remains on par with methods such as DPM and FED, which also introduce substantial architectural modifications to ViT. Importantly, during inference, only the student branch of OGFR is utilized, leading to a significant reduction in inference time. Our method exhibits a considerably lower inference latency than DPM while achieving performance comparable to TransReID. The comparative results, evaluated across multiple metrics, demonstrate that our method achieves superior performance within an appropriate level of complexity. This highlights the effectiveness and practical applicability of OGFR as a highly competitive solution for occluded Re-ID tasks.

\section{Conclusion}
In this work, we propose a novel Occlusion-Guided Feature Purification Learning via Reinforced Knowledge Distillation (OGFR), a teacher-student distillation framework to handle diverse occlusion scenarios. It leverages a variant Vision Transformer to adaptively learn occlusion-aware robust representations, allowing it to effectively generalize to diverse and unseen occlusion scenarios. Furthermore, our Feature Erasing and Purification (FEP) Module adaptively identifies and replaces low-quality patch tokens within holistic images, thereby mitigating feature contamination issues and improving the extraction of identity-related discriminative features. By employing reinforced knowledge distillation, OGFR successfully transfers purified holistic knowledge from the holistic branch to the occluded branch. Extensive experiments on six popular occluded and holistic datasets demonstrate the superiority and effectiveness of our proposed method.

\bibliographystyle{IEEEtran}
\bibliography{cite}

\vfill

\end{document}